\documentclass[10pt,twocolumn,letterpaper]{article}

\usepackage{wacv}              

\usepackage[accsupp]{axessibility}
\usepackage{graphicx}
\usepackage{amsmath,amssymb,amsfonts,bm}
\usepackage{array}
\usepackage{textcomp}
\usepackage{stfloats}
\usepackage{multicol,multirow}
\usepackage{url}
\usepackage{verbatim}
\usepackage{graphicx}
\usepackage{graphicx}
\usepackage{caption}
\usepackage{url}            
\usepackage{booktabs}       
\usepackage{nicefrac}       
\usepackage{microtype}      
\usepackage{cuted} 
\usepackage{xcolor}
\usepackage{float}
\usepackage{wrapfig}
\usepackage[ruled,linesnumbered]{algorithm2e} 
\usepackage{algpseudocode}
\usepackage[pagebackref,breaklinks,pagebackref,colorlinks]{hyperref}

\def\wacvPaperID{1133} 
\def\confName{WACV}
\def\confYear{2025}

\newcommand{\floor}[1]{\lfloor #1 \rfloor}
\newlength{\tabep} 
\newlength\savedwidth
\newcommand\whline{\noalign{\global\savedwidth
		\arrayrulewidth\global\arrayrulewidth \tabep}
	\hline \noalign{\global\arrayrulewidth
		\savedwidth}
}

\algrenewcommand{\algorithmiccomment}[1]{{\textcolor{gray}{//\small{#1}}}}

\usepackage[capitalize]{cleveref}
\crefname{section}{Sec.}{Secs.}
\Crefname{section}{Section}{Sections}
\Crefname{table}{Table}{Tables}
\crefname{table}{Tab.}{Tabs.}

\begin{document}

	\title{Learning deep illumination-robust features from multispectral filter array images}
\author{Anis Amziane\\
	Luxembourg Institute of Science and Technology (LIST), ENVISION Unit\\
	4362 Esch-sur-Alzette, Luxembourg\\
	{\tt\small anis.amziane@list.lu}
}

	\maketitle
		
	\begin{strip}
		\centering
		\includegraphics[width=.85\textwidth,height=5.41cm]{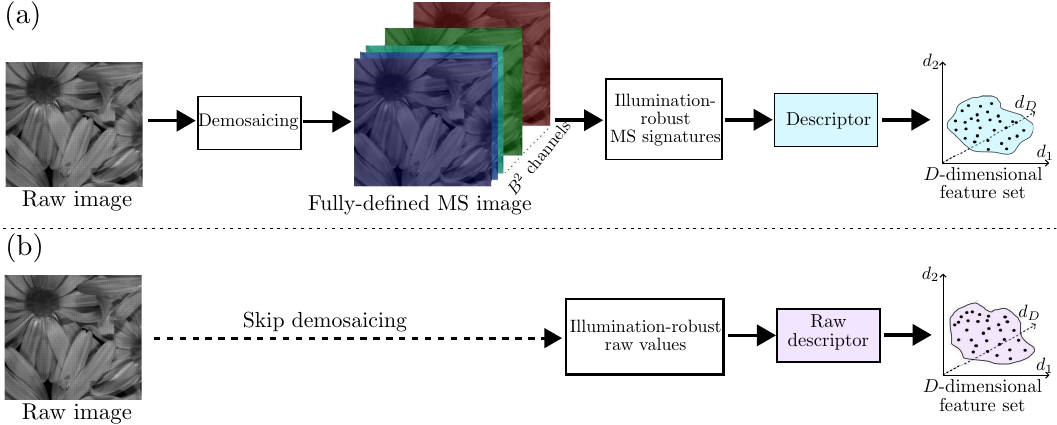}
		\captionof{figure}{Fully-defined (a) vs. proposed (b) raw feature extraction pipeline. (a) requires estimating the fully-defined
			image first before post-processing and feature extraction, while (b) directly exploits the acquired raw image.}
		\label{graphical_abstract}
	\end{strip}
	
	\begin{abstract}
		Multispectral (MS) snapshot cameras equipped with a MS filter array (MSFA), capture multiple spectral bands in a single shot, resulting in a raw mosaic image where each pixel holds only one channel value. The fully-defined MS image is estimated from the raw one through \textit{demosaicing}, which inevitably introduces spatio-spectral artifacts. Moreover, training on fully-defined MS images can be computationally intensive, particularly with deep neural networks (DNNs), and may result in features lacking discrimination power due to suboptimal learning of spatio-spectral interactions. Furthermore, outdoor MS image acquisition occurs under varying lighting conditions, leading to illumination-dependent features. This paper presents an original approach to learn discriminant and illumination-robust features directly from raw images. It involves: \textit{raw spectral constancy} to mitigate the impact of illumination, \textit{MSFA-preserving} transformations suited for raw image augmentation to train DNNs on diverse raw textures, and \textit{raw-mixing} to capture discriminant spatio-spectral interactions in raw images. 
		Experiments on MS image classification show that our approach outperforms both handcrafted and recent deep learning-based methods, while also requiring significantly less computational effort. The source code is available at https://github.com/AnisAmziane/RawTexture.
		
	\end{abstract}

	\section{Introduction}
\begin{figure}[ht!]
		\hspace*{-0.5cm}
		\includegraphics[width=9.1cm,height=3.cm]{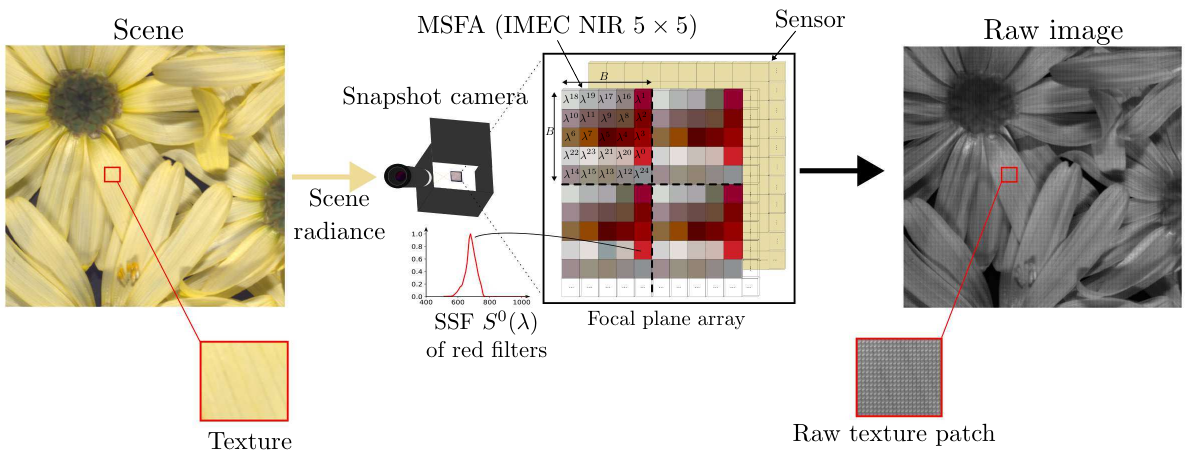}
		\caption{MSFA-based snapshot imaging: shown is the IMEC 
			NIR $5\times5$ MSFA ($\lambda^{b}\in\{678\,nm,~\dots,~960\,nm\}$, $b\in[\![0, 24]\!]$), composed of $B^{2}=25$ optical filters arranged in a $5\times5$ square basic pattern. It samples the scene radiance according the SSF $S^{b}(\lambda)$ of each filter, and provides a raw image whose pixels contain a single channel value.}
		\label{snapshot_process}
	\end{figure}

	Much of the recent success in several pattern recognition applications, such as crop/weed recognition~\cite{amziane_qcav2021,wendel_inproceedings_2016} and texture classification~\cite{mihoubi_josaa_2018,amziane_prl2023}, is thanks in part to the advancement of cameras and specifically to the emergence of compact "snapshot" MS cameras~\cite{hagen_oe_2013}. These cameras rely on an MSFA laid over the sensor to acquire several spectral
	bands in one shot and provide a raw mosaic image, in which a single channel value is available at each pixel. Each filter of the MSFA samples the scene radiance in a given spectral band according to its spectral sensitivity function (SSF), so that each pixel of the
	acquired raw image represents a single band (see Fig.~\ref{snapshot_process}). The fully-defined MS image is estimated from the acquired raw values by demosaicing. To mitigate the impact of outdoor illumination variation on scene spectral signatures, the demosaiced image may undergo an additional post-processing step, such as white balancing or reflectance estimation~\cite{amziane_sensors2021}.
	With their ability to capture information across several narrow spectral bands, MS cameras transcend the limitations of the trichromatic human vision.
	Concurrently, DNNs have shown remarkable ability to decipher complex patterns within images. The fusion of MS imaging and deep learning empowered researchers to push pattern recognition into realms previously inaccessible~\cite{zhi_cvpr2019,huang_lsa2022}. Nevertheless, despite the potential of this fusion, significant challenges remain. Indeed, DNNs and particularly convolutional neural networks (CNNs), require substantial computational resources for training on MS images. Moreover, the MS demosaicing step also increases computational costs 
	and generates spatio-spectral artifacts, which alters the spatio-spectral representation learning. Furthermore, in outdoor settings, MS images are acquired under different illumination conditions, which makes learned features illumination-dependent. To enjoy the benefits of recent deep learning methods and snapshot-based MS imaging, it is important to adopt a tailored strategy that leverages the strengths of both paradigms. For this purpose, we propose learning from raw images as a substitute to the classical approach that learns from fully-defined MS images.\\
	\textbf{Contribution.~}This paper deals with MS texture classification under different lighting conditions. Specifically, we propose an original and computationally efficient approach to extract illumination-robust and discriminant features directly from raw images. Our contributions are summarized as follows:
	\begin{itemize}
		\item We propose \textit{raw spectral constancy} by extending a statistical-based color constancy method to the raw domain to make raw values illumination-robust. 
		\item To diversify raw textures and to ensure sufficient images for training deep models, we propose \textit{MSFA-preserving} augmentations that maintain the structure of the MSFA basic pattern during image augmentation.
		\item  We propose a hybrid architecture called \textit{RawMixer} that exploits a raw convolutional mixer and a transformer encoder for learning discriminant deep spatio-spectral interactions in raw images.   
	\end{itemize}
	
	\section{Related works}
	To classify MS images captured by a single-sensor snapshot camera that samples $B^2$ bands through an MSFA, the typical approach involves estimating fully-defined ($B^2$-channel) images from the acquired raw ones through demosaicing~\cite{mihoubi_ci_2017,antonucci_dsw2019}, followed by feature extraction~\cite{mirhashemi_mva2018,porebski_prl2022, mihoubi_josaa_2018}. Various studies have focused on developing sophisticated handcrafted~\cite{lee_tip2012,mihoubi_josaa_2018, chu_ieeetip2021} and deep learning-based~\cite{hong_itgrs2022,liang_grsl2022,sun_tgrs2022} methods tailored for MS images. However, handcrafted methods often yield high-dimensional features that are not easily interpretable~\cite{dubley_tip2016,mihoubi_josaa_2018}, and deep learning-based approaches are greedy in computation time and memory, making learning from high spectral resolution images challenging. To address these issues, some studies propose to skip the demosaicing step and learn from raw images directly~\cite{amziane_prl2023,mihoubi_josaa_2018,omar_sensors2023}.
	
	\subsection{Learning from raw images}
	If a descriptor effectively processes a raw image, it has the potential to achieve 
	comparable or superior classification performances compared to demosaiced images because demosaicing generates artifacts that may compromise the quality of feature representation. Mihoubi et al. \cite{mihoubi_josaa_2018} proposed a method based on the local binary pattern (LBP) operator to compute texture features from raw images. The method analyzes the raw image with respect to the MSFA basic pattern and its band arrangement 
	to build a texture descriptor by concatenating $B^2$ LBP histograms. Chen et al. \cite{chen_isprs2022} proposed histogram of oriented mosaic gradients (HOMG) to extract spatio-spectral features from raw images. First- and second-order horizontal and vertical gradient operators are designed to capture the spatial and spectral information. Then, the raw image is convolved with the designed gradient operators, resulting in spatio-spectral gradient maps. From these maps, four 31-dimensional histograms are constructed based on the magnitude and orientation of the gradients. The final 124-dimensional feature vector is obtained by concatenating these four histograms. Amziane et al. \cite{amziane_prl2023} proposed a CNN called MSFA-Net for feature extraction from raw reflectance images, guided by the MSFA basic pattern. Their model is composed of three different convolutional layers where the first layer captures spatio-spectral interactions between the channel values in each basic pattern, and two max-pooling layers in between. The produced feature maps are averaged channel-wise then fed into a fully-connected layer to provide the final 128-dimensional feature vector. In \cite{omar_sensors2023}, a modified VGG-11 model pre-trained on ImageNet dataset is re-trained with raw images and used for raw texture feature extraction under different illumination conditions. Oppositely to~\cite{amziane_prl2023}, this approach applies a CNN directly to raw images without proper consideration of the basic pattern which may lead to potential inconsistency in spatio-spectral representation learning. Askary et al.~\cite{askary_sensors2024}~utilized the Places-CNN network, pre-trained on RGB images from the Places dataset. Three pseudo-RGB images are created from each raw image, and each is processed individually by a separate pre-trained Places-CNN. The final prediction is determined by combining the softmax probabilities from all three networks and selecting the class with the highest score. While this method exploits raw images, it does not directly learn from them. Moreover, it disrupts the original spatio-spectral scene structure present in each raw image. Furthermore, a separate network for each pseudo RGB image is required, leading to high computational demands and becomes impractical as the number of spectral bands grows (e.g., 8 networks for raw images obtained by a 25-band MSFA).
	In contrast to the aforementioned methods, our approach first mitigates the impact of illumination on raw images before being processed by the network. Additionally, we apply specific augmentations that maintain the MSFA basic pattern integrity throughout the augmentation process. Finally, we design a hybrid deep model guided by the MSFA basic pattern to learn consistent and discriminant deep features directly from raw images.

	

	\section{Method} \label{methods}
	In this Section, we describe our method for illumination-robust learning from raw images. First, we describe our approach to white-balance (or correct) the raw radiance values to achieve raw spectral constancy. Then, we present our MSFA-preserving augmentations. Finally, we introduce RawMixer architecture for learning from the raw images.
	\subsection{Raw spectral constancy}\label{raw_constancy}
	A single-sensor MS camera fitted with an MSFA provides a raw image
	$I^{\text{raw}}$ of size $X\times X$, where $X=m\cdot B$ is a multiple of the MSFA basic pattern width $B$, with $m$ an integer value representing the scaling factor for the image size relative to the basic pattern size $B$. A single band $b \in \{0, . . . , B^{2}-1\}$ is associated with each pixel $p$ according to the MSFA. Under the Lambertian surface assumption and a global illumination source, $I^{\text{raw}}$ can be described as a function mainly dependent on three physical factors: the illumination relative spectral power distribution (RSPD) $L(\lambda)$, the surface element $s$ spectral reflectance $R_{s}(\lambda)$, and the sensor SSFs~ $\{S^{b}(\lambda)\}_{b=0}^{B^{2}-1}$. By discarding the optical attenuation of the lens, the image integration time, and the quantization function, the radiance measured at pixel $p$ of the raw image $I^{\text{raw}}$ can be then simply expressed as:
	\begin{equation}
		\label{MS image formation with Snapscan}
		I_{{p}}^{\text{raw}} =  \int_{\Omega} L(\lambda)\cdot R_{s}(\lambda)\cdot S^{\mathrm{MSFA}(p)}(\lambda) \:d\lambda  \text{,}
	\end{equation}
	where $\mathrm{MSFA}(p)$ is the function that associates each pixel $p$ with the corresponding band index $b$, and $\Omega$ is the working spectral domain. Under the narrow-band assumption~\cite{finlayson_color_2001}, this model can be further simplified (like in~\cite{barron_iccv2015}) as:
	\begin{equation}
		\label{simplified_MS_model}
		I_{{p}}^{\text{raw}} =  	\tilde{I}_{{p}}^{\text{raw}} \cdot L^{\mathrm{MSFA}(p)}   \text{,}
	\end{equation}
	where $\tilde{I}_{{p}}^{\text{raw}}$ is the white-balanced (WB) raw value at pixel $p$, and $L^{\mathrm{MSFA}(p)}$ is the illumination value associated to the band indexed by the function $\mathrm{MSFA(p)}$. This model assumes that raw spectral constancy can be achieved by independently adjusting the raw values for each channel. Through the application of a periodic pixel-unshuffling operator $\mathcal{PU}:~{\mathbb{R}^{m\cdot B\times m\cdot B}}\mapsto \mathbb{R}^{m\times m \times B^2}$, $I^{\text{raw}}$ can be transformed into a $B^2$-channel image $\textbf{O}^{(B^2)}$ by rearranging its pixels as:
	\begin{equation}
		\label{pixel_unshuffle}
		O_{p(x, y)}^{\mathrm{MSFA}(p(x\cdot B, y\cdot B))} =\mathcal{PU}(I^{\text{raw}},x,y,B) =  I^{\text{raw}}_{p(x\cdot B, y\cdot B)}  \text{,}
	\end{equation}
	where  $(x, y) \in [\![0, m-1]\!]^2$ represents the pixel coordinates in the unshuffled image. Equation.~\ref{simplified_MS_model} can be rewritten as the element-wise Hadamard product of the WB pixel-unshuffled raw image $\tilde{\textbf{O}}^{(B^2)}= \mathcal{PU}(\tilde{I}^{\text{raw}})$ and the illumination vector $\textbf{L}^{(B^2)}$ as:
	\begin{equation}
		\label{simplified_MS_model_2}
		\textbf{O}^{(B^2)} =  	\tilde{\textbf{O}}^{(B^2)}\odot\textbf{L}^{(B^2)}   \text{.}
	\end{equation}
	The goal is to estimate
	$\textbf{L}^{(B^2)}$, so as to recover $\tilde{\textbf{O}}^{(B^2)}$, given $\textbf{O}^{B^2}$. For this purpose,  we propose \textit{Max-Raw}, an extension to the raw domain of the foundational color constancy \textit{Max-RGB}~\cite{gijsenij_tip2011} method. Raw values are WB pixel-wise as:
	\begin{equation}
		\label{wb_max_raw}
		\tilde{O}_{p}^{b} = \dfrac{O_{p}^{b}}{\underset{p}{\text{max}}~\underset{5\times5}{\text{median}\{O^b}\}}   \text{,}
	\end{equation}
	where the denominator is the estimated illumination associated to the $b$-th channel, and $\underset{5\times5}{\text{median}}$\{\} is a $5\times 5$ median filter to avoid corrupted and noisy pixels. Lastly, $\tilde{\textbf{O}}^{(B^2)} $ is reorganized (shuffled) back to $\tilde{I}^{\mathrm{raw}}$ using the shuffling operator $\mathcal{PS}:~{\mathbb{R}^{m\times m\times B^2}}\mapsto \mathbb{R}^{m \cdot B \times m \cdot B}$: 
	\begin{equation}
		\label{pixel_shuffle}
		\tilde{I}^{\mathrm{raw}}_{p(x\cdot B,y\cdot B)} =\mathcal{PS}(\tilde{\textbf{O}}^{(B^2)}, x, y)= \tilde{O}^{\mathrm{MSFA}(p(x\cdot B,y\cdot B))}_{p(x, y)} \text{.}
	\end{equation}
	
	
	\subsection{MSFA-preserving augmentation}\label{msfa_preserving}
\begin{figure*}[ht!]
		\centering
		\includegraphics[width=17.cm,height=5.cm]{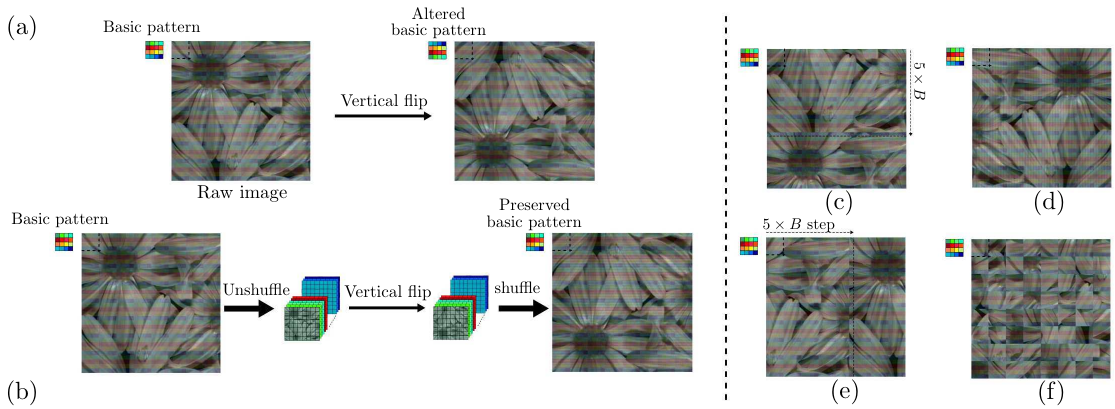}
		\caption{Example of MSFA-preserving augmentations. The direct vertical flip in (a) provides an augmented raw image with a flipped basic pattern while our proposed vertical flip (b) preserves its structure, crucial for learning from raw images. (c) and (e) are translations along the y and x-axis, respectively. (d) is the horizontal flip and (f) the texture remodeling augmentation. Basic patterns are enlarged for better visualization.}
		\label{msfa_aug}
	\end{figure*}

	Traditional image augmentation techniques like rotation, cropping, and flipping are not directly applicable to raw images as they alter the structure of the basic pattern (see Fig.~\ref{msfa_aug}(a)).  In \cite{jiaming_cvpr219}, a method for Bayer pattern preservation during horizontal and vertical flips has been introduced. This involves column and/or row cropping after the augmentation to obtain the correct patterning of the raw image. However this approach is unsuitable for MSFAs with no redundant bands. To address this, we introduce seven augmentations tailored for MSFAs without redundant bands, applied to the unshuffled raw image instead of the original raw image. These transformations include MSFA-preserving vertical and horizontal flips, translation along x and y-axis, and texture remodeling. Additionally, we introduce Gaussian noise (directly to the raw image, with $\mu=0$ and $\sigma=0.25$), as a common artifact in raw image acquisition, and optical distortion. This distortion bends the image (barrel effects), causing straight lines to appear curved, much like the effect of a fish-eye lens. Horizontal and vertical flips are performed by unshuffling the raw image (see Eq.~\eqref{pixel_unshuffle}), applying the flip, and then shuffling it back (see Eq.~\eqref{pixel_shuffle}) to preserve the band arrangement. Translations are performed with step sizes that are multiples of the basic pattern width $B$ along the x and y-axis. Texture remodeling involves randomly replacing raw sub-patches matching the basic MSFA pattern with sub-patches from different locations within the raw image.
	
	\subsection{Raw feature extraction}\label{raw_fe}
\begin{figure*}[ht!]
		\centering
		\includegraphics[width=18.3cm,height=8.02cm]{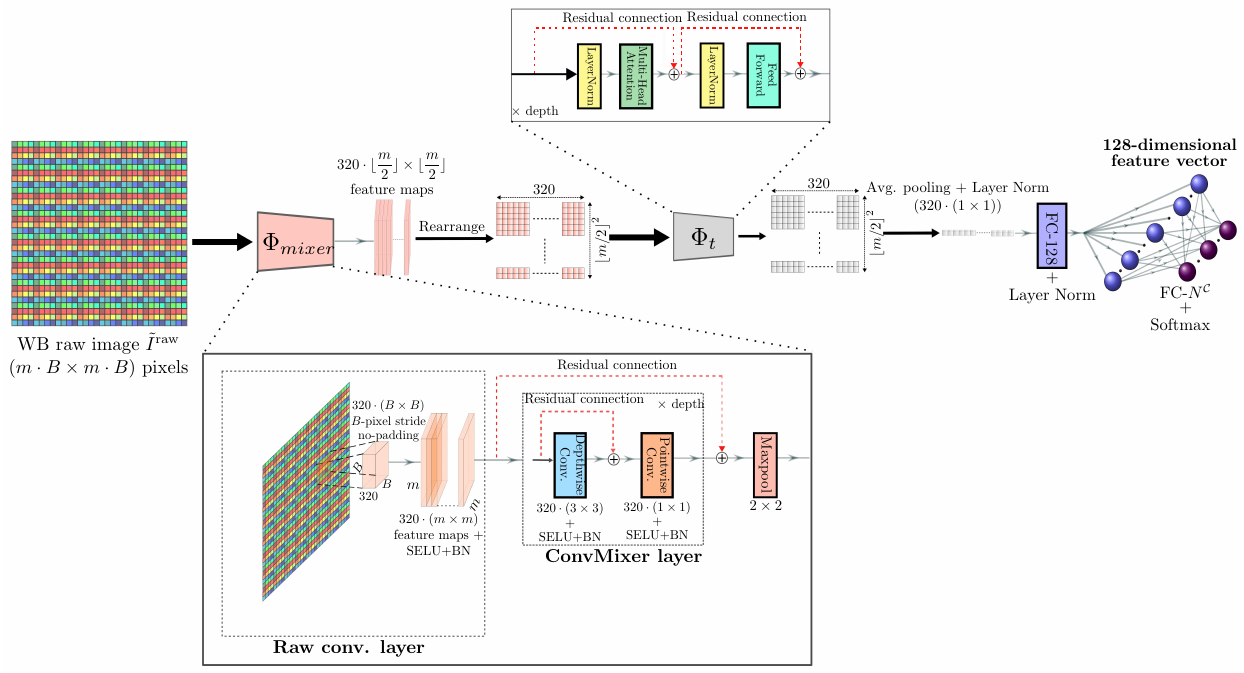}
		\caption{RawMixer architecture. $\Phi_{mixer}$ learns deep spatio-spectral interactions guided by 
			the MSFA basic pattern. $\Phi_t$ is the (positional encoding free) transformer encoder that takes in the $320\cdot (\lfloor{\frac{m}{2}\rfloor}\times \lfloor{\frac{m}{2}\rfloor})$ feature maps provided by $\Phi_{mixer}$ reshaped as $\lfloor{\frac{m}{2}\rfloor}^2$ tokens $\times$ $320$ features. It learns another embedding through self-attention and feed-forward layers. The depth of the ConvMixer and transformer encoders is set to 2 in our experiments.
			 SELU: scaled exponential linear unit, BN: batch normalization, FC: fully-connected layer. Filter depths (1 in raw conv. layer, 320 for ConvMixer layer) are not shown for sake of clarity.}
		\label{rawmixer_fig}
	\end{figure*}

	We introduce RawMixer (see Fig.~\ref{rawmixer_fig}), a hybrid network that combines the strengths of both transformers and CNNs to capture discriminant spatio-spectral interactions in raw images. The WB input raw image $\tilde{I}^{\mathrm{raw}}$ (see Eqs.~\eqref{wb_max_raw} and \eqref{pixel_shuffle}) of size $m\cdot B \times m\cdot B$ pixels, is first fed to the raw ConvMixer $\Phi_{mixer}$ for spatio-spectral feature extraction. Its raw convolutional layer is designed to guide the feature extraction according to the MSFA basic pattern. It uses $320$ 
	convolutional kernels $\{H_n\}_{n=0}^{319}$ of size $B\times B$ and depth~1, with a stride of $B$ pixels along both spatial dimensions and no padding. The $B$-pixel stride ensures that each kernel coefficient remains associated with the same MSFA band throughout all convolutions.
	This first layer learns spatio-spectral interactions among 
	channel values within each raw basic patch that matches the basic MSFA pattern. The convolution between $\tilde{I}^{\mathrm{raw}}$ and a kernel $H_{n}$, $n\in[\![0, 319]\!]$, is defined at each pixel $(x,y) \in[\![0, m-1]\!]^{2}$ as:
	\begin{equation}\label{msfa_conv}
		F_{n}(x,y) = \sum_{i =0}^{B-1}\sum_{j = 0}^{B-1} H_{n}(i,j)\cdot \tilde{I}^{\mathrm{raw}}(B\cdot x+i,B\cdot y+j).
	\end{equation}
	The resulting feature maps $\{F_n\}_{n=0}^{319}$ of size $m\times m$ pixels are processed through the mixing block, employing depth-wise and point-wise convolutions to blend feature values across spatial locations and channels. Subsequently, the resulting feature maps are added to the previous ones through a residual connection before undergoing a $2\times 2$ maxpooling operation to yield a feature map of size $320\cdot(\floor{\frac{m}{2}}\times \floor{\frac{m}{2}})$, that is reshaped and fed to the transformer encoder for attention computation. The resulting matrix is average pooled token-wise (each pixel is a token) to provide a 320-dimensional feature vector. This vector is then passed through a fully-connected layer to add non-linearity and reduce its dimensionality to obtain the final 128-dimensional texture feature vector, which serves as input to the softmax layer.
	\section{Experiments}\label{experiments}
	In this section, we investigate the effectiveness of our approach by conducting experiments on MS texture classification. For this 
	purpose, we simulate raw radiance images based on the HyTexila~\cite{khan_s_2018} and SpecTex~\cite{mirhashemi_mva2018} datasets.
	\subsection{MSFAs}
\begin{figure}
		\hspace*{-0.8cm}
		\begin{tabular}{ccc}
			\begin{tabular}[b]{c}
				\includegraphics[width=2.6cm,height=2.4cm]{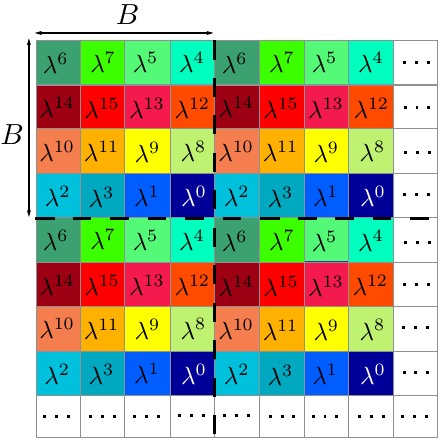}\\
				(a)
			\end{tabular}	
			&
			\begin{tabular}[b]{c}
				\includegraphics[width=2.8cm,height=2.5cm]{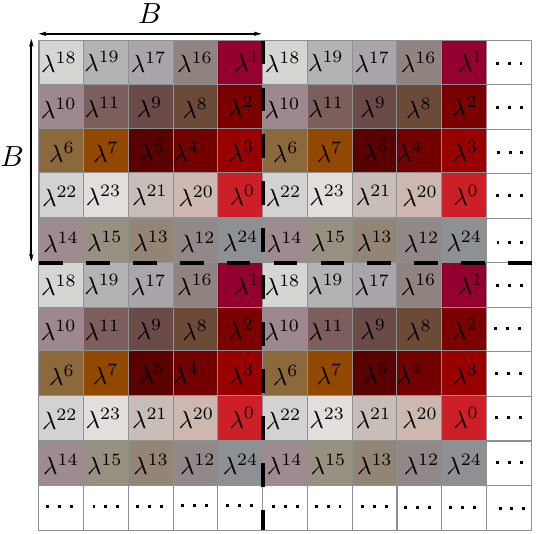}\\
				(b)
			\end{tabular}	
			&
			\begin{tabular}[b]{c}
				\includegraphics[width=2.1cm,height=2.1cm]{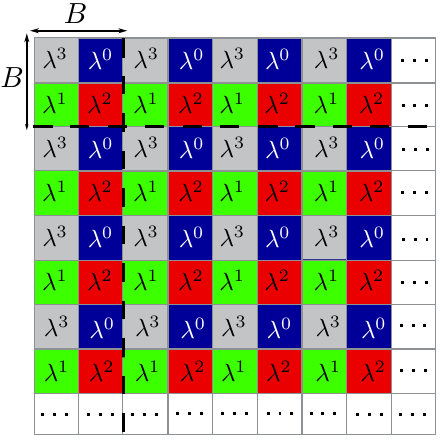}\\
				(c) 
			\end{tabular}	
		\end{tabular}	
		\caption{Considered MSFAs: (a) IMEC VIS $4\times4$ ($\lambda^{b}\in\{469\,nm,~\dots,~633\,nm\}$, $b\in[\![0, 15]\!]$),  
			(b) NIR $5\times5$ ($\lambda^{b}\in\{678\,nm,~\dots,~960\,nm\}$, $b\in[\![0, 24]\!]$), (c)  and (c) VIS-NIR
			$2\times2$ ($\lambda^{b}\in\{465\,nm,~\dots,~811\,nm\}$, $b\in[\![0, 3]\!]$).}
		\label{consiered_MSFAs}
	\end{figure}

	We consider three MSFAs (see Fig.~\ref{consiered_MSFAs}) that are defined by repetition of a $B\times B$ basic
	pattern that samples $B^2$ different bands. We follow the MSFA arrangements in the VIS domain ($B=4$), NIR domain ($B=5$), and VIS-NIR\footnote{\href{https://www.imechyperspectral.com/sites/default/files/2021-02/SNAPSHOT\%20RGB\%2BNIR\%20MULTISPECTRAL\%20IMAGING\%20CAMERA.pdf}{IMEC VIS-NIR $2\times 2$ MSFA}\label{footnote2}} domain ($B=2$) of three snapshot cameras from IMEC~\cite{geelen_spie_2014}. 

	\subsection{Datasets}
	\subsubsection{HyTexila}
\begin{figure}
		\includegraphics[width=8.cm,height=4.2cm]{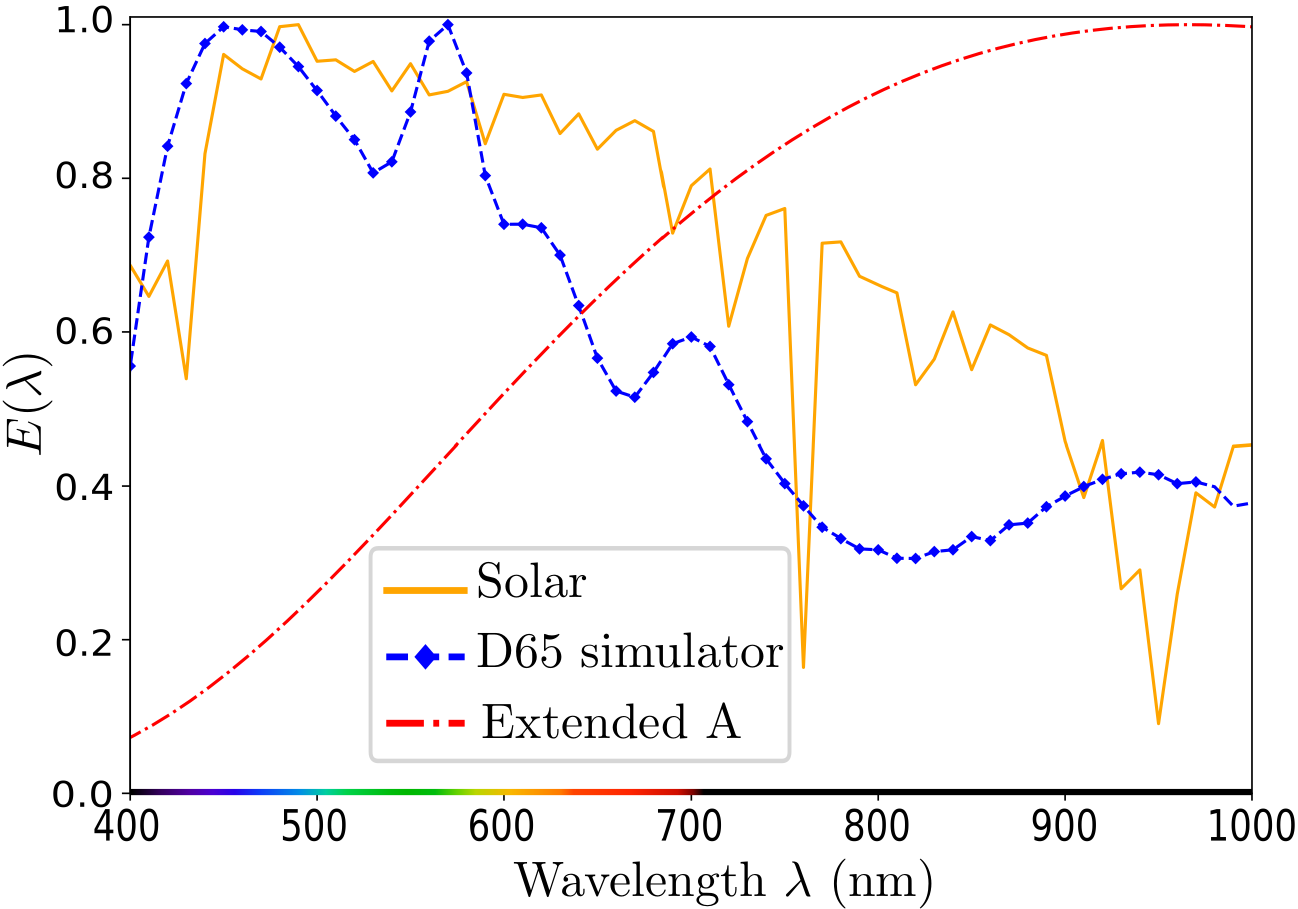}
		\caption{RSPDs of the considered illuminants.}
		
		\label{illuminants}
	\end{figure}

	It contains 112 MS reflectance images each of which can be regarded as a distinctive class. A reflectance image $\mathbf{R}^{(K)}=\{R^{k}\}_{k=0}^{K-1}$ has $K=186$ channels of size $1024 \times 1024$ pixels, and each channel $R^{k}$ is associated to a spectral band of central wavelength $\lambda^{k}\in[405.37\,\text{nm},\,995.83\,\text{nm}]$. We simulate the radiance of HyTexiLa scenes under different illuminants adapted to multispectral imaging, as these illuminants are also defined along the NIR domain of the spectrum (see Fig.~\ref{illuminants}), namely: Solar, D65 simulator, and extended A illuminants~\cite{thomas_sensors2016}. 
	Then, each $K$-channel radiance image is transformed into a $B^2$-channel one ($B^2\in\{4,16,25\}$) by 
	selecting the channels whose associated wavelengths are the closest to the SSF centers of IMEC snapshot cameras.  Finally, we simulate raw radiance images by spatio-spectrally sub-sampling the fully-defined $B^2$-channel images according to the $4\times4$, $5\times5$, or $2\times2$ MSFA.
	\subsubsection{SpecTex}
	It contains 60 MS reflectance images of textures where each can be regarded as a single class. Each image has 77 channels in the range $[400\,\text{nm},780\,\text{nm}]$ of size $640\times 640$ pixels. We simulate raw radiance images following the same procedure used for HyTexila dataset.
	\subsection{Classification protocol}\label{classif_protocol}
	\textbf{Patch extraction}.~For HyTexila, each image is horizontally split into two sub-images. To ensure that the first pixel (top-left corner) aligns with the first band of the MSFA, the sub-images are sized at either $508\times1024$ (for $B=2$ and $B=4$) or $510\times1024$ (for $B=5$) pixels. Non-overlapping square patches of width $X = n \cdot B$ are extracted from them, such as those associated to the second sub-image are used as test patches. We extract square patches of sizes $124^2$ and $64^2$ pixels (for $B=2$ and $B=4$), or $125^2$ and $65^2$ pixels ($B=5$). We perform MSFA-preserving augmentations (see Sec.~\ref{msfa_preserving}) to introduce some texture variations and to ensure enough patches to train the 
	networks. We train on patches associated to D65 simulator and test on patches associated to extended A and Solar illuminants. To make the texture recognition more challenging, we perform horizontal/vertical flips, translation (with a step of $a\times B$ pixels, $a=\lfloor0.6\cdot n\rfloor$) along y-axis, and texture remodeling augmentations only on learning patches, while the remaining ones (Gaussian noise, horizontal translation, and optical distortion) are performed on the test patches. The number $N_{l}$ of training patches is then $N_{l} \approx 27.4 \cdot 10^3$ \text{for} $X \in \{64, 65\}$, and $N_{l} \approx 8.9 \cdot 10^3$ \text{for} $X \in  \{124, 125\}$, and the number $N_{t}$ of test patches is $N_{t} \approx 21.95 \cdot 10^3$ \text{for} $X \in \{64, 65\}$, and $N_{t} \approx 7.1 \cdot 10^3$ \text{for} $X \in  \{124, 125\}$. \\
	For SpexTex, we also horizontally split each image into two sub-images of size $320\times320$ pixels. Note that IMEC25 MSFA is not considered here because SpecTex images are defined only in the VIS domain, which corresponds to IMEC4 and IMEC16 spectral bands. Square patches of sizes $124^2$ and $64^2$ pixels (for $B=2$ and $B=4$) are extracted and augmented following the same procedure used for HyTexila. 
	The number $N_{l}$ of training patches is then $N_{l} \approx 48 \cdot 10^2$ \text{for} $X = 64$, and $N_{l} \approx 38.4 \cdot 10^2$ \text{for} $X =  124$, and the number $N_{t}$ of test patches is $N_{t} \approx 21.95 \cdot 10^3$ \text{for} $X  = 64$, and $N_{t} \approx 960$ \text{for} $X = 124$.\\
	\textbf{Descriptors}. As hand-crafted features, we compute histograms of LBP operators, including the marginal LBP \cite{mihoubi_josaa_2018}, local angular patterns (LAP) \cite{cusano_iciap2015}, LBP-LCC \cite{cusano_josaa2014}, and the M-LBP \cite{mihoubi_josaa_2018} that operates on raw patches.
	As deep learning-based features, we consider MSFA-Net~\cite{amziane_prl2023} and the model (here
	called VGG11$_{\text{raw}}$) used in \cite{omar_sensors2023} to learn from raw images. To learn from demosaiced images, we opt for several state-of-the-art descriptors. The 18-layer (ResNet18\footnote{\href{https://www.paperswithcode.com/model/resnet}{https://www.paperswithcode.com/model/resnet}\label{footnote1}}) model for feature extraction using deep residual learning~\cite{he_cvpr2016}. The HSI-Mixer~\cite{liang_grsl2022} and the SpectralFormer~\cite{hong_itgrs2022} models, specifically designed for hyperspectral image classification. The Vision Transformer (ViT)~\cite{dosovitskiy_iclr2021} and the hybrid Vision Conformer (Vi-Conformer)~\cite{iwana_icdar} models to incorporate attention during feature extraction.\\ 
	\textbf{Feature extraction}. 
	To extract features from fully-defined images, we first estimate the $B^2$-channel MS images by demosaicing using the state-of-the-art alternating steepest descent (ASD) method \cite{antonucci_dsw2019}, initialized with a first guess provided by the spectral difference (SD) method. Then, we perform WB using the max-spectral~\cite{khan_josaa_2017} method before performing training/feature extraction. To extract features from raw images, we perform WB by Max-Raw method (see Eqs.~\eqref{wb_max_raw} and \eqref{pixel_shuffle}), followed by training/feature extraction. For transformer and hybrid models, we use the learned class tokens as the feature vector, while for CNN-based descriptors, we use the features of the last fully-connected layer before the softmax one.\\
	\textbf{Training of DNNs}.
	We train using 95\% of the training patches, the remaining 5\% are used for validation. We use the AdamW weight
	optimizer for all models. The loss function to be minimized is
	the cross-entropy loss. Optimization is conducted over 30 epochs with a constant learning rate of $2e$-$4$ and weight
	decay of $1e$-$5$ (training parameters were determined experimentally). The batch size is set to 128.
	\subsection{Results and discussion}
\begin{figure*}[ht!]
\hspace*{-0.64cm}
	\begin{tabular}{cc}

		\includegraphics[width=8.7cm,height=5.66cm]{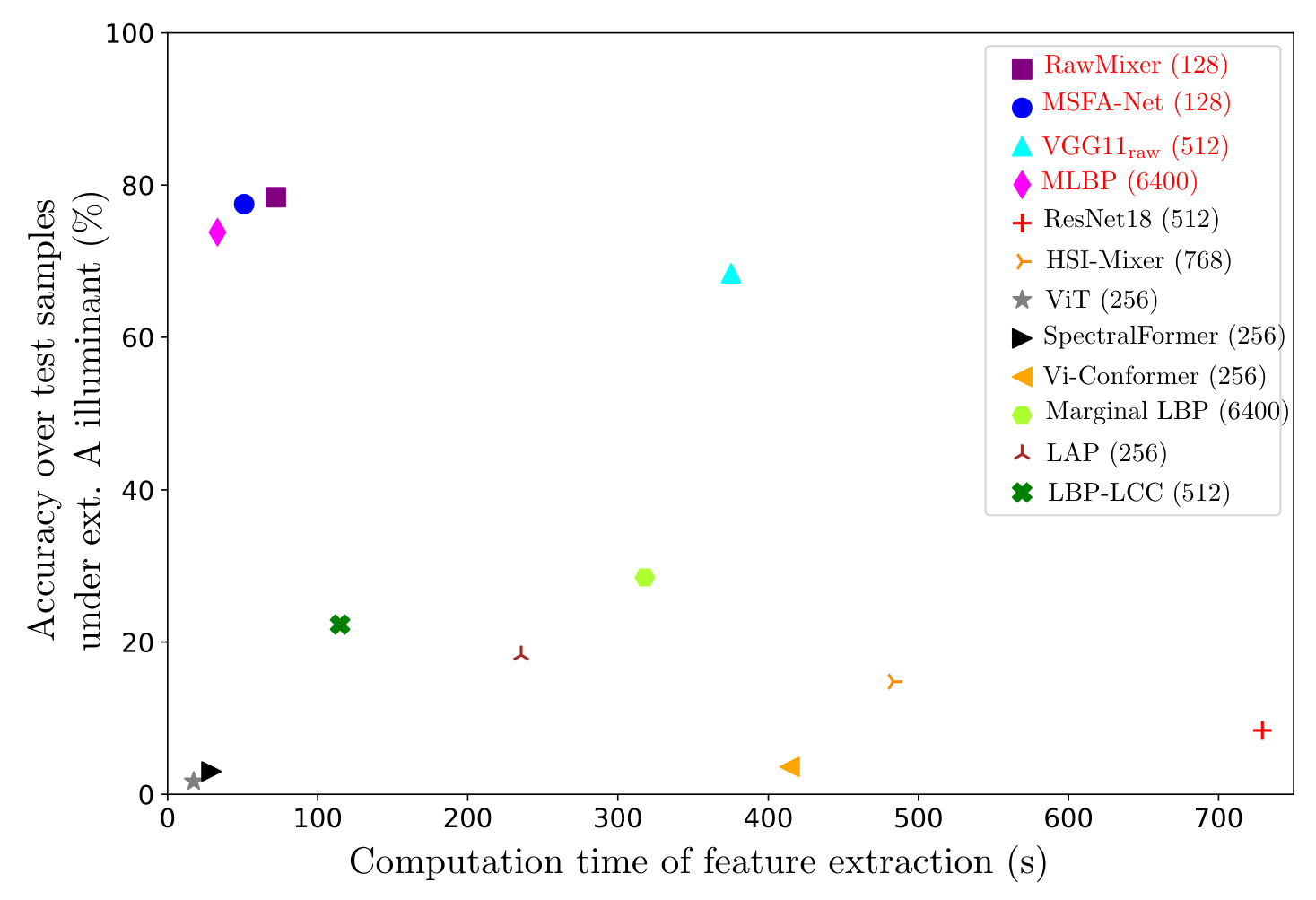} &
		\includegraphics[width=8.7cm,height=5.66cm]{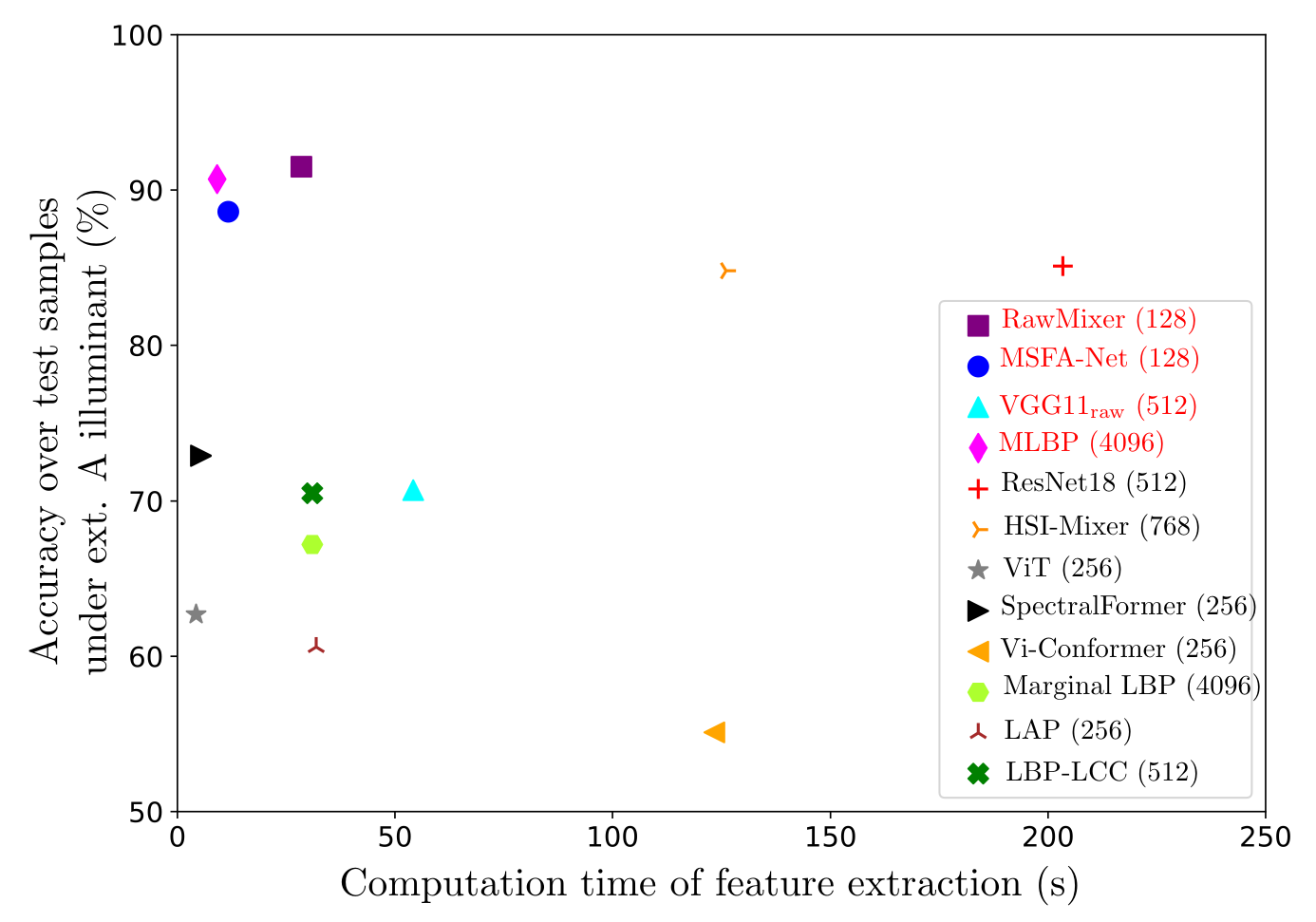}\\
		(a) IMEC $5\times5\mid65\times65\mid N_{l} \approx 27.4 \cdot 10^3$ & (b) IMEC 		$4\times4\mid124\times124\mid N_{l} \approx 38.4 \cdot 10^2$
	\end{tabular}
	\caption{1-NN classification accuracy vs. computation time of feature extraction from the learning patches associated to  HyTexila (a) and SpecTex (b) datasets. Raw descriptors are highlighted in red, and numbers in parentheses indicate the dimension of each feature. Demosaicing computation time (e.g., $\approx$41s/image for HyTexila images with IMEC $5\times5$ MSFA) is not considered.}
	\label{feature_extraction_cost}
\end{figure*}

	\begin{table*}[ht!]
		\caption{1-NN classification accuracy (\%) on HyTexiLa (a) and SpecTex (b) datasets with features extracted from either raw or demosaiced test patches. Training/testing are performed according to the protocol described in Sec.~\ref{classif_protocol}. The best result in each column is shown as bold and second best as italics. Superscript $^*$ refers to IMEC $4\times 4$, $^\dag$ to IMEC $5\times 5$, and  $^\ddag$ to IMEC $2\times 2$. Note that IMEC $5\times 5$ is not considered for SpecTex dataset since it's not defined in the NIR domain.}
		\begin{subtable}{\textwidth}
			\centering
			\label{HyTexiLa results}
			\makebox[\textwidth]{	\resizebox{.93\textwidth}{!}{	\begin{tabular}{|c|lr|c|c|c|c||c|c|c|c||c|c|c|c|} 
						\hline
						\multirow{3}{*}{\begin{tabular}[c]{@{}c@{}}Input\\patches\end{tabular}} & \multicolumn{2}{l|}{\multirow{3}{*}{\begin{tabular}[c]{@{}l@{}}\\\\Descriptor~~~~~~~~~~~~~~~~~~~~~~~~~~~~~~~~~~~~~~~~~~~~~~~~Feat. size\end{tabular}}} & \multicolumn{4}{c||}{IMEC $4\times4$$^*$ (VIS)}                                                        & \multicolumn{4}{c||}{IMEC $5\times5$$^\dag$ (NIR)}                                   & \multicolumn{4}{c|}{IMEC $2\times2$$^\ddag$ (VIS-NIR)}                 \\ 
						\cline{4-15}
						& \multicolumn{2}{c|}{}                                                                                                                            & \multicolumn{2}{c|}{$64\times 64$}             & \multicolumn{2}{c||}{$124\times 124$}          & \multicolumn{2}{c|}{$65\times 65$} & \multicolumn{2}{c||}{$125\times125$} & \multicolumn{2}{c|}{$64\times 64$} & \multicolumn{2}{c|}{$124\times 124$}  \\ 
						\cline{4-15}
						& \multicolumn{2}{l|}{}                                                                                                                            & A & Solar                 & A                     & Solar                 & A & Solar     & A & Solar                           & A &Solar&A &Solar                        \\ 
						\whline
						\multirow{4}{*}{Raw MSFA}                                                   & RawMixer (ours)     & 128$^{*,\dag, \ddag}$                                                                                                                    & \textbf{59.7}  & 83.9 & \textbf{64.8} & 86.0 & \textbf{80.6} & \textbf{78.4}  & \textbf{82.6}  & \textbf{80.2}                                 &                            47.4        &      74.6    & \textbf{56.9} &  76.5      \\ 
						\cline{2-15}
						& MSFA-Net~\cite{amziane_prl2023}       & 128$^{*,\dag, \ddag}$                                                                                                                    & \textit{52.6} & 81.5 &  52.2 & 81.9 & \textit{79.0}  & \textit{77.5} & \textit{78.8}  & \textit{75.6}                                &                   \textit{47.9}                 & 72.8 & \textit{52.5}  &  73.7           \\ 
						\cline{2-15}
						& VGG11$_{\mathrm{raw}}$~\cite{omar_sensors2023}       & 512$^{*,\dag, \ddag}$                                                                                                                    & 52.0 & 80.1 &  \textit{54.7} & 80.8 &  69.5 & 68.4 & 72.4  &  71.7                              &                                    \textbf{52.6} &    73.0 &\textit{52.5} &71.6             \\ 
						\cline{2-15}
						& M-LBP~\cite{mihoubi_josaa_2018}          & 4096$^*$/6400$^\dag$/1024$^\ddag$                                                                                                            &   49.4  &  79.3 &  49.0   & 80.7  & 77.2 & 73.8 &  77.7 & {74.5} &   23.9 & 55.8 &  25.3    &  58.7                \\ 
						\whline
						\multirow{8}{*}{Demosaiced}                                             & ResNet18\footref{footnote1}       & 512$^{*,\dag, \ddag}$                                                                                                                    &   38.5                     &  \textbf{93.1}                     &   27.7                    &            \textbf{88.7}          &           50.9             &    8.4    & 56.3  &          33.5                         &   24.4 & \textit{85.7} &        29.0                       &    \textbf{86.2}              \\ 
						\cline{2-15}
						& HSI-Mixer~\cite{liang_grsl2022}      & 768$^{*,\dag, \ddag}$                                                                                                                   &   17.3                     &        \textit{88.6}               &    23.8                   &          \textit{87.0}             &                 37.4       &   14.8        & 39.7  &         18.4                        &     21.7        &  75.7  &        16.7           &    57.1              \\ 
						\cline{2-15}
						& ViT~\cite{dosovitskiy_iclr2021}            & 256$^{*,\dag, \ddag}$                                                                                                                    &     29.2                   &   87.0                    &        21.8               &            67.5           &  31.0                      &  1.7         & 24.7  &     6.2                            &         25.3       & 84.7 &            17.4       &      72.6            \\ 
						\cline{2-15}
						& SpectralFormer~\cite{hong_itgrs2022} & 256$^{*,\dag,\ddag}$                                                                                                                    &   33.4                     &    86.9                   &         21.7              &        73.4               &   38.0                     &  3.0         & 27.7  &  8.0                               &  25.3     & \textbf{86.1} &            21.1                 &   \textit{80.1}               \\ 
						\cline{2-15}
						& Vi-Conformer~\cite{iwana_icdar}   & 256$^{*,\dag, \ddag}$                                                                                                                    &   31.2                     &       82.9                &  16.7                     &  73.1                    &  30.7                       &  3.6         & 25.3  &  5.7                               &       24.5          &   85.0   &     18.0         &   75.8               \\ 
						\cline{2-15}
						& Marginal LBP~\cite{mihoubi_josaa_2018}   & 4096$^*$/6400$^\dag$/1024$^\ddag$                                                                                                            &  10.0   &  41.9   &  10.5  & 49.4                      &       26.6                 &      28.5     &   29.4&       33.8                          &             3.0                       &   34.0   & 2.2 &    34.1        \\
						\cline{2-15}
						& LAP~\cite{cusano_iciap2015}            & 256$^{*,\dag, \ddag}$      
						& 9.4  & 28.2 & 9.4 & 35.5 & 16.3                       &  18.3         &  20.9 &        22.7                         &             7.23             &  20.9    &  8.2  &  25.1                \\ 
						\cline{2-15}
						& LBP-LCC~\cite{cusano_josaa2014}   & 512$^{*,\dag, \ddag}$                                                                                                            &    20.0   & 36.4  &  22.1  & 44.8 &              29.5          &    22.3       &  34.9 &                 25.7                &               6.7                   &    29.1    & 2.2&    32.7      \\
						\hline
			\end{tabular}}}
			\subcaption{HyTexila dataset}
		\end{subtable}\\
		\begin{subtable}{\textwidth}
			\centering
			\makebox[\textwidth]{	\resizebox{0.79\textwidth}{!}{
							\begin{tabular}{|c|lr|c|c|c|c||c|c|c|c|c|} 
						\hline
						\multirow{3}{*}{\begin{tabular}[c]{@{}c@{}}Input\\patches\end{tabular}} & \multicolumn{2}{l|}{\multirow{3}{*}{\begin{tabular}[c]{@{}l@{}}\\\\Descriptor~~~~~~~~~~~~~~~~~~~~~~~~~~~~~~~~~~~~~~~~~~~~~~~~Feat. size\end{tabular}}} & \multicolumn{4}{c||}{IMEC $4\times4$$^*$ (VIS)}                                                        & \multicolumn{4}{c|}{IMEC $2\times2$$^\ddag$ (VIS-NIR)}                 \\ 
						\cline{4-11}
						& \multicolumn{2}{c|}{}                                                                                                                            & \multicolumn{2}{c|}{$64\times 64$}             & \multicolumn{2}{c||}{$124\times 124$}          & \multicolumn{2}{c|}{$64\times 64$} & \multicolumn{2}{c|}{$124\times 124$}  \\ 
						\cline{4-11}
						& \multicolumn{2}{l|}{}                                                                                                                            & A & Solar                 & A                     & Solar                 & A & Solar     & A & Solar                        \\ 
						\whline
						\multirow{4}{*}{Raw MSFA}                                                   & RawMixer (ours)     & 128$^{*,\ddag}$                                                                                                                    & \textbf{83.1} &  {91.4}& \textbf{82.6} &\textbf{91.5}  &          \textbf{77.9}                &   \textbf{84.6}       & \textbf{79.5}  &   \textbf{85.2}     \\ 
						\cline{2-11}
						& MSFA-Net~\cite{amziane_prl2023}       & 128$^{*,\ddag}$                                                                                                                    & \textit{79.2} & {89.0} &  \textit{77.8} & 88.6 &                   \textit{73.5}       & \textit{83.5} & \textit{77.2}  &       \textit{84.5}      \\ 
						\cline{2-11}
						& VGG11$_{\mathrm{raw}}$~\cite{omar_sensors2023}       & 512$^{*,\ddag}$                                                                                                                    & 74.9 & 82.5 & 64.0  & 70.7 &                              72.1       & 79.6  &67.4 &       73.4      \\ 
						\cline{2-11}
						& M-LBP~\cite{mihoubi_josaa_2018}          & 4096$^*$/1024$^\ddag$                                                                                                            &  73.2   &  89.3 &  73.4  & \textit{90.7}  &  42.3  & 69.5 &    42.2  &   69.5               \\ 
						\whline
						\multirow{8}{*}{Demosaiced}                                             & ResNet18\footref{footnote1}       & 512$^{*,\ddag}$                                                                                                                    &           29.8            &  \textit{92.1}                     & 21.5                     &         85.1            &  63.6  & 59.1 &     28.8                         &       63.6          \\ 
						\cline{2-11}
						& HSI-Mixer~\cite{liang_grsl2022}      & 768$^{*,\ddag}$                                                                                                                   &        25.2                &      89.1                 &  26.9                     &       84.8                &        46.3     &  58.0  &    30.6               &    46.3            \\ 
						\cline{2-11}
						& ViT~\cite{dosovitskiy_iclr2021}            & 256$^{*,\ddag}$                                                                                                                    &   21.1                    &   83.9                   &        15.7               &         62.7             &      50.5          & 59.7 &  13.6                 &   49.1               \\ 
						\cline{2-11}
						& SpectralFormer~\cite{hong_itgrs2022} & 256$^{*,\ddag}$                                                                                                                    &          32.0             &         \textbf{92.7}            &    23.3                  &        72.9               &   49.2  & 63.1 &      15.7                       &     49.2            \\ 
						\cline{2-11}
						& Vi-Conformer~\cite{iwana_icdar}   & 256$^{*,\ddag}$                                                                                                                    &        22.2                &        82.3              &     19.2                  &          55.1           &         49.1       &  60.5   &     13.2        &       50.5           \\ 
						\cline{2-11}
						& Marginal LBP~\cite{mihoubi_josaa_2018}   & 4096$^*$/1024$^\ddag$                                                                                                            &  50.1   &  72.0 &  41.7  & 67.2 &  16.0 &    52.7    & 16.2 &   47.3        \\
						\cline{2-11}
						& LAP~\cite{cusano_iciap2015}            & 256$^{*,\ddag}$      
						& 31.8  & 46.3 & 34.6& 60.6 &  26.7 &     38.9 &   29.9  &    39.4             \\ 
						\cline{2-11}
						& LBP-LCC~\cite{cusano_josaa2014}   & 512$^{*,\ddag}$                                                                                                            &   39.7   & 60.3 & 47.7  & 70.5 &                                 40.2  &     57.9 & 38.2 &     52.1     \\
						\hline
			\end{tabular}}
		}
			\subcaption{SpecTex dataset}
		\end{subtable}
		\label{classification results}
	\end{table*}
	Table~\ref{classification results} shows the classification results obtained for each feature with 1-nearest neighbor (NN) classifier coupled with the Euclidean distance.
	On HyTexila dataset (see Tab.~\ref{classification results}(a)), M-LBP outperforms the other
	LBP-based descriptors since it considers spatio-spectral correlations
	more effectively within the raw image, while avoiding the demosaicing step that affects the texture representation. Texture features extracted by deep learning-based 
	descriptors from demosaiced patches outperform LBP, LAP, and LBP-LCC features. Their performance is strongly related to both MSFA bands and illuminant. Although WB is performed, the computed features (histograms) have poor discrimination power especially with patches simulated under illuminant A. 
	ResNet18, HSI-Mixer, ViT, SpectralFormer, and Vi-Conformer are outperformed by raw descriptors utilizing IMEC $5 \times 5$ patches because they struggle to capture 
	discriminant spatio-spectral interactions from the 25-channel demosaiced patches. However,  their performance is significantly enhanced under illuminant A relative to Solar illuminant. This phenomenon can be attributed to the higher RSPD of illuminant A in the NIR bands sampled by IMEC $5\times5$ filters.
	ResNet18 and SpectralFormer are ranked first three times and one time among the 12 tested cases, respectively.
	 They perform better under Solar illuminant in the VIS and VIS-NIR domains, likely because textures are more prominently highlighted in these conditions. However, their performance declines compared to RawMixer and MSFA-Net, possibly due to the artifacts introduced by the demosaicing procedure, which can disrupt texture representation. Additionally, capturing spatio-spectral correlations from demosaiced images is challenging, even with advanced learning paradigms. Replacing the 2D convolutional kernels in some layers with 3D kernels could improve their performance, as 3D kernels capture both spatial and spectral information simultaneously. Our proposed approach is ranked first seven times among the 12 tested cases, followed by MSFA-Net that is ranked second seven times, then VGG11$_{\mathrm{raw}}$, ranked first once and second twice. Unlike RawMixer and MSFANet, VGG11$_{\mathrm{raw}}$ struggles to learn consistent spatio-spectral correlations because its first layer is not guided by the MSFA basic pattern. On the SpecTex dataset (see Tab.~\ref{classification results}(b)), RawMixer ranks first in seven out of the eight tested cases, followed by MSFA-Net, which ranks second in six cases. While VGG11$_{\mathrm{raw}}$ performs better than M-LBP with IMEC $2\times2$, M-LBP features outperform those of VGG11$_{\mathrm{raw}}$ with IMEC $4\times4$, likely because M-LBP captures more spatio-spectral channel correlations with this MSFA. Additionally, M-LBP surpasses the other tested LBP-based descriptors. Similar to the HyTexila results, ResNet18, SpectralFormer, and HSI-Mixer deliver strong performances in the VIS domain under Solar illuminant, likely because textures are better highlighted with IMEC $4\times4$ under Solar conditions.
RawMixer exhibits robustness to unseen augmentations and benefits from raw spectral constancy, making its features robust to illumination variation and thus generalizable across different illuminants and textures.
	Furthermore, learning from raw images by RawMixer provides comparable or superior performance to learning from $B^2$-channel demosaiced images at significantly reduced computation costs (see Fig.~\ref{feature_extraction_cost}).
	Its performance is minimally impacted by the MSFA basic pattern size, with similar performance observed under different illuminants (e.g., IMEC $2\times2$ and IMEC $4\times4$ under A illuminant, and IMEC $2\times2$ and IMEC $5\times5$ under Solar illuminant), hinting at the potential extension to the color domain by utilizing different CFAs like the Bayer CFA~\cite{lukac_tce2005}.

	\subsection{Ablation study}
	\subsubsection{Effect of raw spectral constancy}
	To highlight the effect of raw spectral constancy on the classification performance when the illumination conditions change, we reproduce the same experiments described in Section~\ref{experiments} using RawMixer but without raw spectral constancy. As we can see from Table~\ref{results_no_wb}, these experiments show a drastic classification performance decline due to illumination dependency of learned features. This highlights the importance of raw spectral constancy to achieve illumination-robust representation learning. 
	\begin{table*}
		\caption{1-NN classification accuracy (\%) on HyTexiLa (first row) and SpecTex (second row) raw test patches achieved by RawMixer features following the protocol described in Sec.~\ref{classif_protocol}. Numbers in parentheses are the results (without ablation) recalled from Table \ref{classification results}. Without raw spectral constancy, learned features become illumination-dependent which drastically alters the texture recognition performance.}
		\label{results_no_wb}
			\centering
			\resizebox{\textwidth}{!}{
				\begin{tabular}{l|c|c|c|c||c|c|c|c||c|c|c|c|} 
					\cline{2-13}
					& \multicolumn{4}{c||}{IMEC $4\times4$ (VIS)}                                                                     & \multicolumn{4}{c||}{IMEC $5\times5$ (NIR)}                                   & \multicolumn{4}{c|}{IMEC $2\times2$ (VIS-NIR)}                                     \\ 
					\cline{2-13}
					& \multicolumn{2}{c|}{$64\times 64$}                   & \multicolumn{2}{c||}{$124\times 124$}                 & \multicolumn{2}{c|}{$65\times 65$} & \multicolumn{2}{c||}{$125\times125$} & \multicolumn{2}{c|}{$64\times 64$} & \multicolumn{2}{c|}{$124\times 124$}  \\ 
					\cline{2-13}
					& A   & Solar                     & A                        & Solar                     & A & Solar     & A   & Solar                         & A & Solar     & A & Solar                             \\ 
					\hline
					\multicolumn{1}{|l|}{HyTexila}& {3.9} (59.7) & {61.2} (83.9) 
					& {3.4} (64.8) & {69.8} (86.0) & {5.6} (80.6)  & {5.2} (78.4)       
					& {4.8} (82.6)     & {2.6} (80.2)     &   {2.9}  (47.4)  &  {17.8}  (74.6) &  {2.7} (56.9)  &   {20.4} (76.5)                           \\ 
					\hline
					\multicolumn{1}{|l|}{SpecTex}  
					 & 8.0 (83.1) & {73.3} (91.4) & 6.0 (82.6)  & {74.2} (91.5)  
					  &	\multicolumn{2}{c|}{-} & \multicolumn{2}{c||}{-}  &
					2.3  (77.9)  &  24.5  (84.6)  &  
					{4.6}  (79.5) &  {27.8}  (85.2)                             \\ 
					\hline     
				\end{tabular}
			}
	\end{table*}

	\subsubsection{Effect of MSFA-preserving augmentation}
	Our learning from raw images approach implies that the first layer of RawMixer learns spatio-spectral interactions following the basic MSFA pattern that has to be the same for all patches to make learning spatio-spectral interactions consistent. To highlight the effect of MSFA pattern preservation during learning, we repeated the experiments from Section~\ref{experiments} with RawMixer using the same augmentations as before but without preserving the MSFA pattern during augmentation.
	 Table~\ref{results_no_msfa_presrving} clearly shows that the performance of features provided by the network significantly drops when it learns from raw images with altered patterns, especially on HyTexila dataset. Our proposed augmentations ensures texture diversification while preserving 
	the MSFA basic pattern, which makes learned features consistent.

\begin{table*}[ht!]
	\caption{1-NN classification accuracy (\%) on HyTexiLa (first row) and SpecTex (second row) raw test patches achieved by RawMixer features following the protocol described in Sec.~\ref{classif_protocol}. Numbers in parentheses are the results (without ablation) recalled from Table \ref{classification results}. In this case, the network learns varying MSFA patterns, leading to inconsistencies and less discriminant features, especially with HyTexiLa patches.}
	\label{results_no_msfa_presrving}
	\centering
	\resizebox{1.\textwidth}{!}{
		\begin{tabular}{l|c|c|c|c||c|c|c|c||c|c|c|c|} 
			\cline{2-13}
			& \multicolumn{4}{c||}{IMEC $4\times4$ (VIS)}                                                                     & \multicolumn{4}{c||}{IMEC $5\times5$ (NIR)}                                   & \multicolumn{4}{c|}{IMEC $2\times2$ (VIS-NIR)}                                     \\ 
			
			\cline{2-13}
			& \multicolumn{2}{c|}{$64\times 64$}                   & \multicolumn{2}{c||}{$124\times 124$}                 & \multicolumn{2}{c|}{$65\times 65$} & \multicolumn{2}{c||}{$125\times125$} & \multicolumn{2}{c|}{$64\times 64$} & \multicolumn{2}{c|}{$124\times 124$}  \\ 
			\cline{2-13}
			& A   & Solar                     & A                        & Solar                     & A & Solar     & A   & Solar                         & A & Solar     & A & Solar                             \\ 
			\hline
			\multicolumn{1}{|l|}{HyTexila}	                                                                         & {43.8} (59.7)    & {67.4} (83.9)	    &
			9.4 (64.8)    & 14.4 (86.0)    & 
			{72.5} (80.6)     &   {71.6} (78.4)       
			&  {75.0} (82.6)   & {75.3} (80.2)   
			&  {43.8} (47.4)     &   {67.4} (74.6)         &  {50.6} (56.9)    & {72.3} (76.5)                                   \\ 
			\hline
			\multicolumn{1}{|l|}{SpecTex}	
			& {81.9} (83.1) & {89.3} (91.4) & {82.4} (82.6) & {89.4} (91.5)  &     	\multicolumn{2}{c|}{-} & \multicolumn{2}{c||}{-}  & {77.0} (77.9)    &    {84.2} (84.6)      &  {78.6} (79.5) &    {84.8} (85.2)      
			\\ 
			\hline     
		\end{tabular}
	}
\end{table*}

	\section{Conclusion}
	
	In this paper, we propose learning illumination-robust features directly from raw images, bypassing the demosaicing step that alters the spatio-spectral scene representation. We introduce raw spectral constancy to minimize illumination effects and a raw image augmentation strategy that preserves the MSFA band arrangement. Finally, we 
	propose a model called RawMixer to capture discriminant spatio-spectral interactions in raw images. Extensive experiments on MS image classification show the relevance of our approach for learning deep, illumination-robust texture features. Future work will focus on enhancing RawMixer by integrating raw spectral constancy into its architecture and developing a specialized encoder for MS image segmentation. Potential applications include outdoor crop/weed recognition.

	
	{\small
		\bibliographystyle{ieee_fullname}
		\bibliography{main.bib}
	}
	
\end{document}